\documentclass[final,1p,numbers]{elsarticle}
\usepackage{bm}
\usepackage{amssymb}
\usepackage{amsmath}

\usepackage{xcolor}

\newcommand{\ie}{i.e.\ }

\begin{document}

\begin{frontmatter}



% \title{Modality Decoupling and Memorizing based Mix-modal Federated Learning for MRI Image Segmentation} %% Article title
\vspace*{-2cm}
\title{Mix-modal Federated Learning for MRI Image Segmentation} %% Article title
\author[5,1,3,4]{Guyue Hu}
\author[5,1,3,4]{Siyuan Song}
\author[5,1,3,4]{Jingpeng Sun}
\author[5,1,3]{Zhe Jin}
\author[5,1,3,4]{Chenglong Li\corref{cor1}}
\cortext[cor1]{Corresponding author: lcl1314@foxmail.com}
% \ead{lcl1314@foxmail.com}
\author[6,1,4]{Jin Tang}

\affiliation[5]{organization={School
of Artificial Intelligence},
                addressline={Anhui University}, 
                postcode={230601}, 
                city={Hefei},
                country={China}}
                
\affiliation[6]{organization={School of Computer Science and Technology},
                addressline={Anhui University}, 
                postcode={230601}, 
                city={Hefei},
                country={China}}

\affiliation[1]{organization={State Key Laboratory of Opto-Electronic Information Acquisition and Protection Technology},
                addressline={Anhui University}, 
                city={Hefei},
%               citysep={}, % Uncomment if no comma needed between city and postcode
                postcode={230601}, 
                country={China}}

\affiliation[3]{organization={Anhui Provincial Key Laboratory of Security Artificial Intelligence},
                addressline={Anhui University}, 
                city={Hefei},
                postcode={230601}, 
                country={China}}

\affiliation[4]{organization={Anhui Provincial Key Laboratory of Multimodal Cognitive Computation}, 
                addressline={Anhui University}, 
                postcode={230601}, 
                city={Hefei},
                country={China}}

% \author{} %% Author name

% \affiliation{organization={},%Department and Organization

%% Abstract
\begin{abstract}
Magnetic resonance imaging (MRI) image segmentation is crucial in diagnosing and treating many diseases, such as brain tumors. Existing MRI image segmentation methods mainly fall into a centralized multimodal paradigm, which is inapplicable in engineering non-centralized mix-modal medical scenarios. In this situation, each distributed client (hospital) processes multiple mixed MRI modalities, and the modality set and image data for each client are diverse, suffering from extensive client-wise modality heterogeneity and data heterogeneity.
In this paper, we first formulate non-centralized mix-modal MRI image segmentation as a new paradigm for federated learning (FL) that involves multiple modalities, called mix-modal federated learning (MixMFL). It distinguishes from existing multimodal federating learning (MulMFL) and cross-modal federating learning (CroMFL) paradigms. Then, we proposed a novel modality decoupling and memorizing mix-modal federated learning framework (MDM-MixMFL) for MRI image segmentation, which is characterized by a modality decoupling strategy and a modality memorizing mechanism. Specifically, the modality decoupling strategy disentangles each modality into modality-tailored and modality-shared information. During mix-modal federated updating, corresponding modality encoders undergo tailored and shared updating, respectively. It facilitates stable and adaptive federating aggregation of heterogeneous data and modalities from distributed clients. Besides, the modality memorizing mechanism stores client-shared modality prototypes dynamically refreshed from every modality-tailored encoder to compensate for incomplete modalities in each local client. It further benefits modality aggregation and fusion processes during mix-modal federated learning. Extensive experiments on two public datasets for MRI image segmentation demonstrate the effectiveness and superiority of our methods. 
\end{abstract}


% \begin{highlights}
% \end{highlights}

%% Keywords

\end{frontmatter}


\section{Introduction}
Medical magnetic resonance imaging (MRI) image segmentation is crucial in clinic diagnosing and treating, such as brain tumors and joint osteoarthritis. Multi-parametric MRI images typically involve four complementary modalities: T1-weighted (T1), contrast-enhanced T1-weighted (T1c), T2-weighted (T2), and T2 fluid attenuation inversion recovery (FLAIR). These modalities usually focus on different aspects of clinical diseases, such as the first two mainly highlighting the tumor core of the brain tumor, while the latter two are good at highlighting the peritumoral edema of the brain tumor \cite{ahamed2023review}. 

In recent years, the rapid development of deep learning has significantly advanced the field of MRI image segmentation. However, most existing MRI image segmentation methods mainly fall into a centralized multimodal paradigm (with the risk of data and privacy leakage), which is inapplicable in non-centralized mix-modal medical scenarios. Specifically, due to the sensitivity and privacy of medical data, it is challenging for different clients (hospitals) to share local data, making traditional centralized model training unsuitable in non-centralized scenarios \cite{huang2022eefed,huang2024federated,zhou2022pflf, wang2025interpretable,zhang2024afl,hu2025federated,li2025enhance,pan2025diverse}. Moreover, in many practical medical scenarios, distributed hospitals often hold different combinations of imaging modalities for various reasons (such as device diversity and modality missing), and their data distribution in each modality is also diverse, causing significant heterogeneity obstacles for decentralized model learning. Thus, it is urgent to develop decentralized mix-modal methods that are appropriate for decentralized mix-modality scenarios.

\begin{figure*}
    \centering
    \includegraphics[width=1\linewidth]{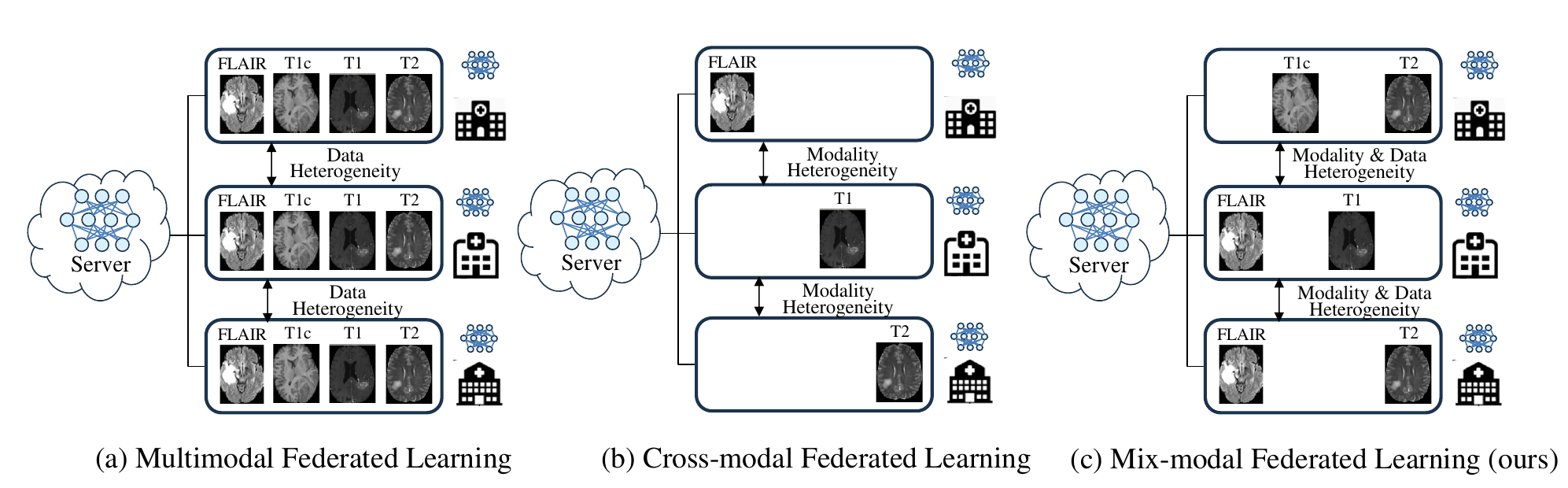}
    \caption{Paradigm comparison of federated learning involving multiple modalities. FLAIR, T1c, T1, and T2 denote different modalities of MRI images, respectively. (a) MulMFL: All multimodal data in the same image modalities but from different data distributions (different hospitals), containing data heterogeneity. (b) CroMFL: Each client holds one different modality from the same data distribution, containing modality heterogeneity. (c) MixMFL: Each client holds multiple mixed modalities and also from different data distributions, containing both modality heterogeneity and data heterogeneity.} 
    \label{overview}
    % \vspace{-0.1cm}
\end{figure*}

To overcome these issues, we first take brain tumor segmentation as an example to formulate practical non-centralized mix-modal MRI image segmentation into a new paradigm of FL involving multiple modalities, called mix-modality federating learning (MixMFL). It distinguishes from existing multimodal federating learning (MulMFL) and cross-modal federating learning (CroMFL) paradigms. As shown in {Fig.}~\ref{overview}, in the MulMFL paradigm \cite{xiong2022unified,ouyang2024admarker}, all clients hold multimodal data in the same image modalities but from different data distributions (different hospitals), containing only data heterogeneity during federated aggregating. In the CroMFL paradigm \cite{yang2024cross,dai2024federated}, each client holds one different modality from the same data distribution, containing only modality heterogeneity during federated aggregating. In contrast, in the proposed  MixMFL ({Fig.}~\ref{overview}~(c)) paradigm, \textit{each client holds multiple mixed modalities and also from different data distributions, containing both modality heterogeneity and data heterogeneity during federated aggregation}. It aims to exploit mixed modalities of heterogeneous data from distributed clients to conduct effective federated learning while ensuring stable and adaptive aggregating processes simultaneously.

Then, we propose a novel modality decoupling and memorizing (MDM) based mix-modal federated learning framework (referred to as MDM-MixMFL) for MRI image segmentation, which is characterized by a modality decoupling strategy and a modality memorizing mechanism. Regarding the modality decoupling strategy, we deploy multiple modality-tailored encoders and one modality-shared encoder for each client (hospital) and decouple each modality into modality-tailored and modality-shared components for tailored and shared federated updating. This strategy enables stable mix-modal modality fusion and data aggregation across distributed mix-modal clients (hospitals) and generates personalized optimal models for each client. Regarding the modality memorizing mechanism, we design a modality memorizing module to memorize, refresh, and retrieve modality prototypes to compensate for incomplete modalities in local clients. This mechanism further benefits modality aggregation and fusion during mix-modal federated learning and improves the segmentation performance in non-centralized mix-modal medical scenarios.

In summary, the main contributions of this paper are as follows:

\begin{itemize}
  \item We formulate the non-centralized mix-modal MRI image segmentation into a new mix-modal federated learning (Mix-MFL) paradigm, which is distinguished from existing multimodal federated learning (MulMFL) and cross-modal federated learning (CroMFL) paradigms.
  \item We propose a novel modality decoupling and memorizing mix-modal federated learning framework (MDM-MixMFL), which enables stable mix-modal modality fusion and data aggregation across distributed mix-modal clients (hospitals) to generate personalized segmentation models for each client.
  \item The proposed modality decoupling strategy adaptively disentangles each modality into modality-tailored and modality-shared information for tailored and shared parameter updating respectively, which facilitates stable and adaptive federated aggregation of heterogeneous data and modalities from distributed clients.
  \item The proposed modality memorizing mechanism buffers modality prototypes dynamically refreshed from every modality-tailored encoder to compensate for incomplete modalities in each local client, which further benefits modality aggregation and fusion during mix-modal federated learning.
  \item Extensive experiments on two public datasets in MRI images for brain tumor segmentation demonstrate the effectiveness and superiority of the proposed framework.
  
\end{itemize}

\section{Related Works}
\subsection{MRI Image Segmentation}
The goal of medical imaging segmentation is to segment patient-specific clinical information to facilitate disease diagnosis and staging, and then guide and monitor interventions, eventually forming personalized treatment \cite{wong2005medical,hu2025dynamic}. Magnetic resonance imaging (MRI) with multimodal modalities allows complementary imaging information from different modalities and provides a comprehensive view of various tissues and lesions. Multimodal MRI performs superior to other single-modal imaging techniques in many disease diagnoses and treatments, such as brain tumors. In recent years, advances in deep learning have significantly improved brain tumor segmentation in multimodal MRI images \cite{zeineldin2022multimodal, ottom2022znet, sun2019brain}. However, these methods are mostly optimized for ideal scenarios, assuming the availability of a complete set of modalities and medical images. In practice, one or more modalities may be unavailable due to image corruption, artifacts, variations in acquisition protocols, allergic reactions to contrast agents, or financial constraints \cite{dai2024federated}. To address this issue, numerous studies have focused on solving the problem of missing or incomplete modalities \cite{shi2023m, liu2024medmap, hu2025missingness}. However, most of these methods rely on centralized training and do not fully address the challenges posed by data privacy restrictions and the inability to share raw data. In this context, our research aims to propose a solution suitable for non-centralized mix-modal environments, enabling multiple clients (hospitals) to collaboratively perform brain tumor segmentation tasks without sharing raw data.

\subsection{Federated Learning with Multiple Modalities}

Existing methods of federated learning that involve multiple modalities can be categorized into Multimodal Federated Learning (MulMFL) and Cross-modal Federated Learning (CroMFL). The MulMFL paradigm assumes that each client possesses the same multiple modalities, with the challenge of locally training effective multimodal models for global aggregation. For example, the MMFed \cite{xiong2022unified}  enhances model performance by fusing complementary information from different modalities through a joint attention mechanism, while the FedHGB \cite{chen2022towards} argues that multiple modality increases data heterogeneity across clients, leading to overfitting and poor generalization. To address this issue, the FedHGB proposes a Hierarchical Gradient Blending mechanism that optimizes modality combinations and balances updates. The CroMFL paradigm allows different clients to have different modalities, aiming to transfer common knowledge across modalities. For example, the FedMEMA \cite{dai2024federated} supports cross-modal information sharing by extracting useful features through an angular margin adjustment and relation-aware calibration mechanism. Similarly, the FedDiff \cite{li2024feddiff} introduces a diffusion-model-based framework that enhances feature extraction and facilitates information exchange. The FedInMM \cite{yu2024robust}  explores effective fusion methods for incomplete multimodal data. 
In addition, recent studies explicitly investigate multimodal federated learning under missing/incomplete-modality settings. 
MFCPL \cite{le2025cross} leverages cross-modal prototypes to compensate for severely missing modalities in federated multimodal learning, while FLISM \cite{orzikulova2024federated} studies federated learning with incomplete sensing modalities and designs strategies to alleviate the performance degradation caused by modality absence. 
These methods primarily focus on missing/incomplete-modality robustness under predefined modality configurations. Additionally, the FedMEMA \cite{dai2024federated} assumes that the server has access to all modalities, while individual clients possess only a single modality.

Although FL involving multiple modalities has achieved significant progress, handling scenarios where distributed clients process different modality combinations (i.e., mix-modality federating learning) is still unexplored. Our discussion highlights the advancements and challenges in multimodal federating learning (MulMFL) and cross-modal federating learning (CroMFL) and points out a promising novel mix-modality federating learning (MixMFL) paradigm needing urgent attention of the federated learning and medical healthcare communities.

\begin{figure*}
    \centering
    \includegraphics[width=1\linewidth]{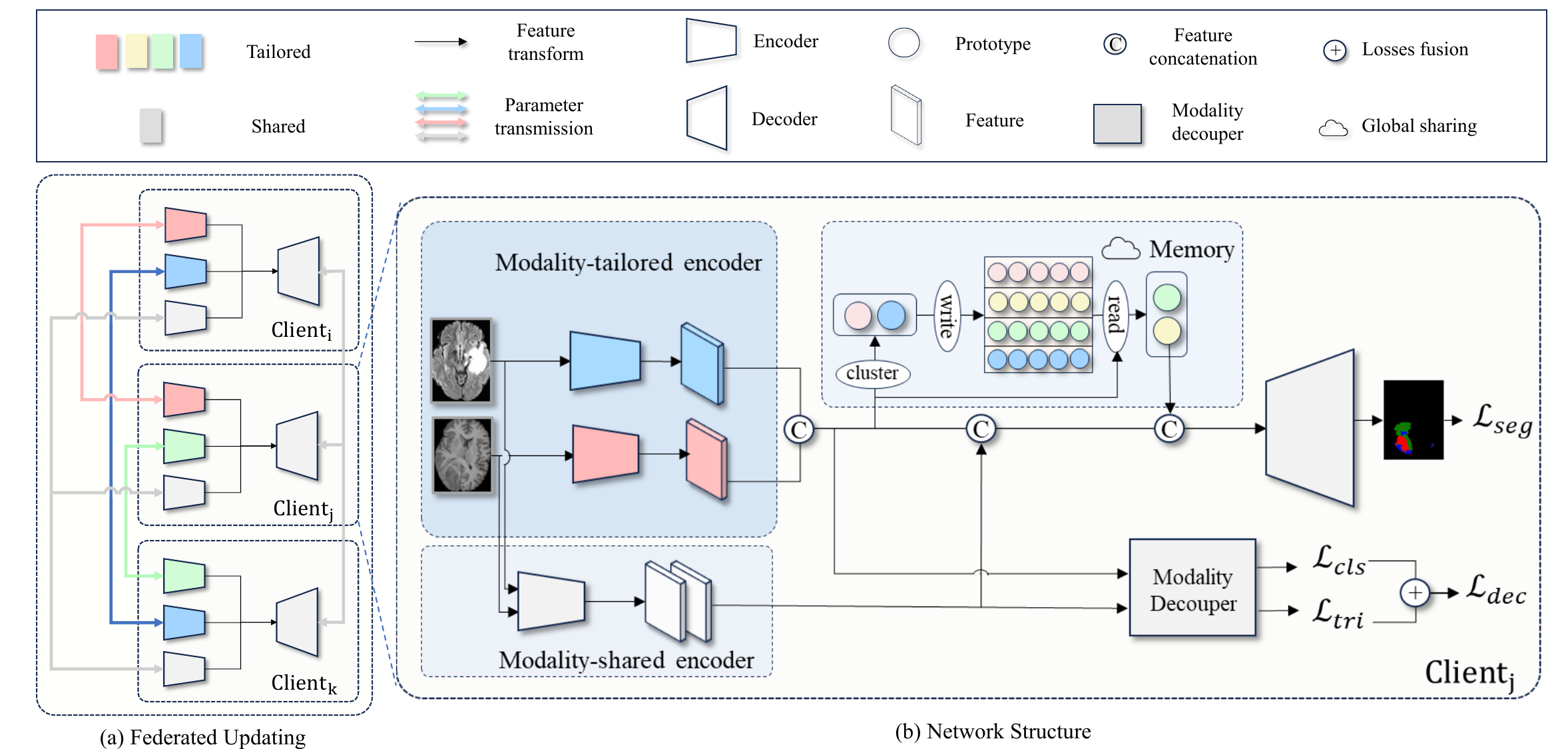}
    \caption{Overview pipeline of the proposed MDM-MixMFL framework. (a) The federated parameter transmission and local feature transform processes among three distributed clients (\textit{i, j, k}). (b) The detailed network structure for every client.}
    \label{framework}
    % \vspace{-0.35cm}
\end{figure*}

\section{Method} 

\subsection{Paradigm Definition of the Mix-modal Federating Learning}
We first take brain tumor segmentation in MRI images as an example to formulate practical non-centralized mix-modal MRI image segmentation into a MixMFL paradigm, which defines a new paradigm for federating learning involving multiple modalities. As shown in {Fig.}~\ref{overview}~(c), assuming MRI images for brain tumor segmentation non-centralized distribute on \(K\) clients (hospitals), denoted as \( \mathcal{D}=\{\mathcal{D}^k, M^k\}_{k=1}^K \), where \( \mathcal{D}^k=\{x_i^k, y_i^k\}_{i=1}^{n_k} \) represents the subset of \( n_k \) samples on the \textit{k}-th client (the total sample number \( N
=\sum_{k=1}^K n_k \)), among which  \( \bm{x}_i^k \) and \( \bm{y}_i^k \) are respectively  the \(i\)-th sample on the \(k\)-th client (\(\bm{x}_i^k \in \mathbb{R}^{1 \times D \times H \times W}\)) and corresponding label. \( M^k \) represents the modality combination on the \(k\)-th client, \ie, ~\( M_k \subsetneq \{T1, T1c, T2, FLAIR\} \). Thus, \textit{each client holds local data in the modality of different proper subsets of all modalities, resulting in client-wise modality heterogeneity and data heterogeneity} (see {Fig.}~\ref{overview}~(c)). The proposed mix-modal federating learning paradigm aims to exploit distributed mix-modal data to collaboratively and adaptively learn a client-tailored (personalized) optimal model for each client while simultaneously overcoming the client-wise heterogeneity from both data and modalities.

\begin{figure}
    \centering
    \includegraphics[width=0.9\linewidth]{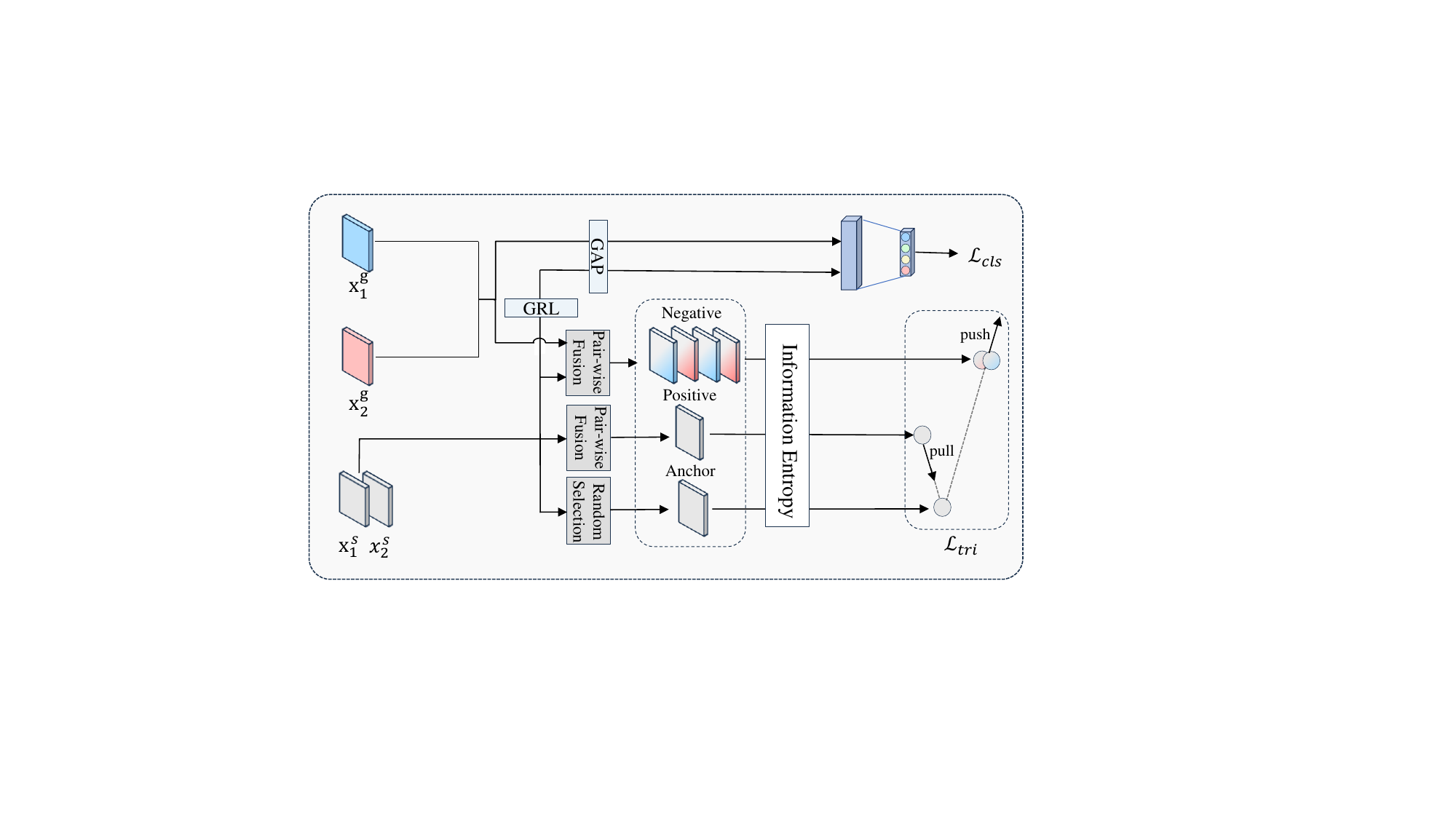}
    \caption{The proposed modality decoupler consists of two main branches accompanied by two different losses.  The above branch performs the modality classification guided by a cross-entropy loss \( \mathcal{L}_{cls} \). The bottom branch performs grouping and merging of modality representations guided by a triplet loss \( \mathcal{L}_{tri} \). GAP denotes the global average pooling layer. GRL denotes the gradient reversal layer, which reverses the parameter gradients and forces the layers below the GRL layer to update along the opposite direction. 
}
    \label{Decouper}
    % \vspace{-0.35cm}
\end{figure}

\subsection{Overview of the Modality Decoupling and Memorizing}
The overview pipeline of the proposed modality decoupling and memorizing (MDM) framework for tackling mix-modal Mix-modal federating learning (MixMFL) is shown in {Fig.}~\ref{framework}~(a). To effectively exploit complementary but diverse modality information in distributed mix-modal scenarios, we deploy multiple modality-tailored encoders and one modality-shared encoder for each client, respectively. The modality-tailored encoder takes in the corresponding modality of images and is designed to encode modality-specific information. It is federally updated according to \textit{those clients having the same modality (modality-tailored encoder)}. In other words, the federal parameter transmitting is only within the same modality and across the clients having the same modality, such as the red, green, and blue lines in {Fig.}~\ref{framework}~(a). The federated process for parameter transmitting, aggregating, and updating for a modality-tailored encoder is similar to the process for conventional FL under the single modality scenarios. In contrast, the modality-shared encoder takes in all modalities of images and is designed to encode modality-invariant information. It is federally updated according to \textit{all modalities in all clients}, such as the grey lines in {Fig.}~\ref{framework}~(a). 

% As shown in {Fig.}~\ref{framework}~(b), to ensure the information in each modality could be effectively disentangled into modality-specific and modality-invariant information, we append an auxiliary modality decoupler guided by a decoupling loss \( \mathcal{L}_{dec} \) during training, which facilitating adaptive modality disentanglement. 

As shown in {Fig.}~\ref{framework}~(b), in the MixMFL setting where different clients observe heterogeneous modality combinations, directly aggregating features across clients may introduce modality-induced inconsistencies and representation instability, a phenomenon widely observed in representation learning with limited memory and evolving data distributions\cite{li2025ckdf}.
To address this issue, we append an auxiliary modality decoupler guided by a decoupling loss \( \mathcal{L}_{dec} \) during training, so that the information in each modality can be explicitly disentangled into modality-specific and modality-invariant components.
This design facilitates more stable and adaptive federated aggregation under heterogeneous modality availability.

As a result, the decoupling strategy could realize adaptively federating aggregation of data and modalities from distributed clients while simultaneously alleviating training instability and slow convergence issues caused by the client-wise heterogeneity from both data and modality. 

Additionally, we design a modality memorizing mechanism to store and refresh modality prototypes for missing modality compensation. Specifically, we extract modality prototypes from the modality-specific representation from the modality-tailored encoder, then store and refresh them in a client-shared memory. Then, we retrieve modality prototypes from the memory to compensate for incomplete modalities in each client based on the existing modalities. As a result, the modality memorizing mechanism further benefits modality aggregation and fusion during mix-modal federated learning.

Eventually, the concatenated representations from the modality-shared encoder, the modality-tailored encoders, and the modality memory are sequentially fed into a Modality-shared decoder to predict the final segmentation mask. The following two sections will introduce the modality decoupling strategy and modality memorizing mechanism in detail.

\begin{figure}
    \centering
    \includegraphics[width=0.6\linewidth]{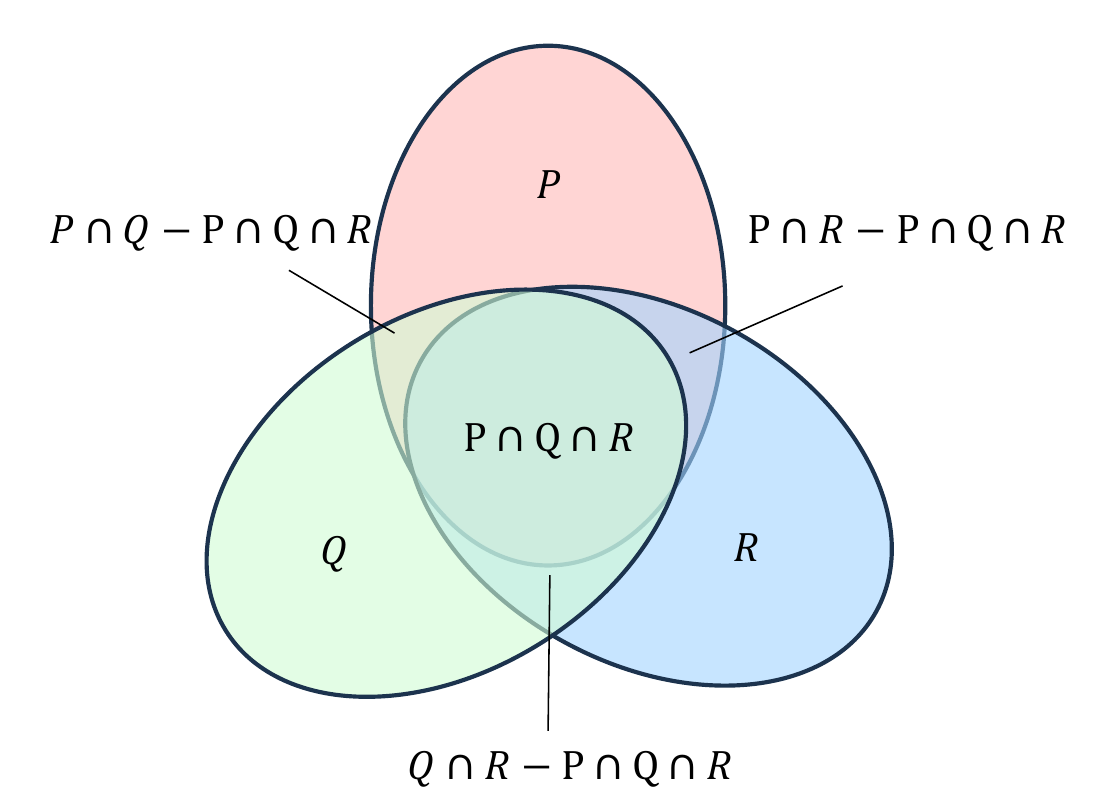}
    \caption{Venn diagram for relation illustration of multiple representation distributions. The symbols \(P\), \(Q\), and \(R\) denote three representation modalities from one local client, respectively.}
    % 特征分布文氏图，P，Q，R表示不同的三种模态的特征分布
    \label{VN}
    % \vspace{-0.35cm}
\end{figure}

\subsection{Modality Decoupling}
As shown in {Fig.}~\ref{Decouper}, the modality decoupler in {Fig.}~\ref{framework}~(b) consists of two main branches accompanied by two different losses. 
These two branches are designed to separately model modality-specific and modality-invariant representations.
By jointly optimizing the two branches, the decoupling module explicitly enforces structured representation learning for subsequent federated aggregation. These two branches and the corresponding loss play a complementary collaboration.

The above branch performs the modality classification guided by a cross-entropy loss \( \mathcal{L}_{cls} \). It takes the output representations of all encoders as input and determines which modality the representations belong to. On the one hand, it encourages the representations generated from the modality-tailored encoders to be easily distinguished from each other, forcing the modality-tailored encoders to be more modality-specific. On the other hand, it also encourages the representations extracted from the modality-shared encoder for different modalities to be hard to distinguish since this portion of the gradient is reversed by the gradient reversal layer (GRL) in {Fig.}~\ref{Decouper}, forcing the modality-shared encoder to be modality-invariant. The modality discrimination loss is defined as follows:

\begin{equation}
\mathcal{L}_{cls} = -\frac{1}{I} \sum_{i=1}^I \sum_{m=1}^M y_{i,m} \log(p_{i,m})
\end{equation}
Where \( I \) is the batchsize, \( y_{i,m} \) represents the ground truth of the \(i\)-th sample for the \(m\)-th modality, and \( p_{i,m} \) represents the classification probability of the \(i\)-th sample for the \(m\)-th modality.

During backpropagation, the Modality-tailored encoder is updated normally, while the modality-shared encoder is updated in the reverse direction. This setup forces each encoder to encode distinct modal representations: each modality-tailored encoder learns to encode representations corresponding to a specific modality, whereas the modality-shared encoder encodes representations that are common to all modalities and don't belong to any specific modality. Eventually, these representations thus could be described as modality-tailored and modality-shared representations, respectively.

\begin{figure*}
    \centering
    \includegraphics[width=1\linewidth]{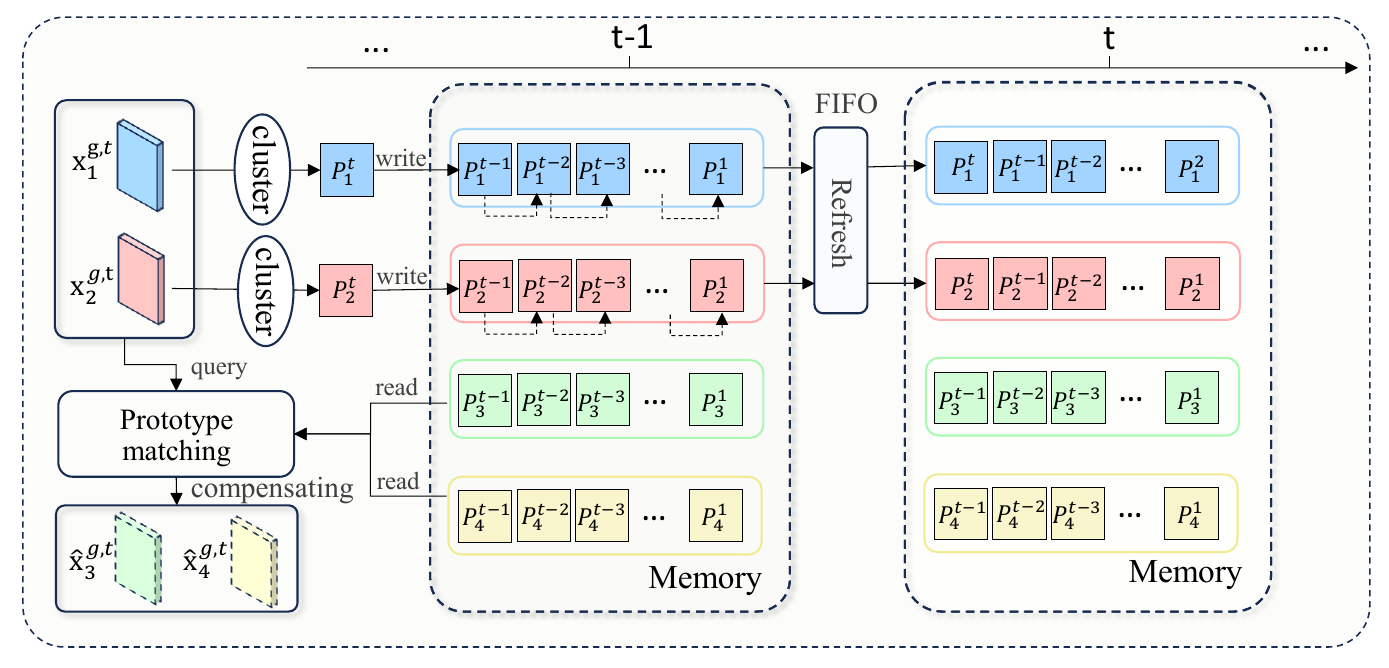}
    \caption{Illustration of the memorizing, refreshing, and retrieving processes of the proposed modality memory mechanism.}
    % Memory模块的示意图
    \label{Memory}
    % \vspace{-0.35cm}
\end{figure*}

Regarding the bottom branch, each modality-tailored encoder generates a modality-tailored representation \(\bm{x}^t\) for the corresponding modality of each sample, and the modality-shared encoder generates a modality-shared representation \(\bm{x}^s\) for all modalities of each sample. We use \(X^g\) and \(X^s\) to denote modality-tailored representation sets and modality-shared representation sets, respectively. Where \(X^g = \{\bm{x}^g_{1},\bm{x}^g_{2}, ...,\bm{x}^g_{m}\}\), \(X^s = \{\bm{x}^s_{1},\bm{x}^s_{2}, ...,\bm{x}^s_{m}\}\), \(m\) is the number of modalities in corresponding client. The Pair-wise Fusion module fuses the representation pairs from \(X^g\) and \(X^s\) via another encoder layer, forming fused representations for each pair. Thus, we construct a representation triplet to optimize the bottom branch: \(Anchor\) (arbitrary modality-shared representation), \(Positive\) (arbitrary fused representation generated from an arbitrary combination consisting of two modality-shared representations, and \(Negative\) (fused representation generated from the arbitrary combination consisting of one modality-shared representation and one modality-specific representation). 
We assume each representation distribution follows a Gaussian distribution, and utilize information entropy to measure the information amount of representation distribution following LaBella’s work \cite{shwartz2017opening}, \ie~\(H(\bm{x}) = \frac{1}{2} \ln(2\pi e \sigma^2)\). Then, the triplet loss encourages modality decoupling by minimizing the entropy distance between the \(Positive\) and \(Anchor\) representations, and maximizing the entropy distance between the \(Negative\) and \(Anchor\) representations, which is formulated as follows:

\begin{align}
&\mathcal{L}_{tri} = \frac{1}{ I}\sum_{i=1}^I  \max(0, 
MaxDis(Anchor_i,Positive_i)  \notag\\&- MinDis(Anchor_i,Negative_i)  + \alpha )
\end{align}
Where \(MaxDis()\) and \(MinDis()\) calculate the maximum and minimum distance regarding information entropy between two representation sets, $\alpha$ is a margin parameter. Eventually, the total loss \(\mathcal{L}\) for our approach consists of conventional segmentation loss \(\mathcal{L}_{seg}\) and decoupling loss \(\mathcal{L}_{dec}\), \ie,

\begin{equation}
\mathcal{L}=\mathcal{L}_{seg}+\mathcal{L}_{dec} = \mathcal{L}_{seg}+\mu\mathcal{L}_{cls}+\gamma\mathcal{L}_{tri}
\end{equation}
Where \(\mu\) and \(\gamma\) are two loss weighing hyperparameters, and \(\mathcal{L}_{seg}\) is the dice loss following Milletari's work \cite{milletari2016v}.

Taking together, the above two branches accompanied by corresponding losses play a complementary collaboration, as shown in {Fig.}~\ref{VN}. The above branch guided by modality classification loss \(\mathcal{L}_{cls}\) encourages identifying modality-specific information but also unreasonable identifies all remaining parts as modality-shared information. Taking a local client with modalities \(P\) and \(Q\) as an example, the parts \(P \cap Q - P \cap Q \cap R\), \(P \cap R - P \cap Q \cap R\), and \(Q \cap R - P \cap Q \cap R\) will be wrongly treated as modality-shared information rather than modality-specific information of any modality. Fortunately, the bottom branch guided by the triple loss \(\mathcal{L}_{tri}\) pulls the \(Anchor\) representations closer to the \(Positive\) representations, making the modality-shared representation distribution gradually approach the intersection of the \(Anchor\)  and the two \(Positive\) representations (\(P \cap Q \)). Similarly, as for clients with modalities \(P\) and \(R\), the modality-shared representation will also approach \(P \cap R \) with a similar process. As a result, two diverse distributions will be generated for clients with the modality \(P\). The aggregating process of federated learning finally turns out to be a global compromise between them, which will result in the final modality-shared representation distribution of \(P\) being the intersection of these three different distributions (\ie, \(P \cap Q\cap R \)). Thus, achieving better decoupling of the modality-shared representations. Similarly, it also pushes the \(Negative\) representation away from the \(Anchor\) representation, which further facilitates the decoupling of the modality-tailored representations on the other hand.

\begin{table}[]
\scriptsize
\centering
\caption{Comparing with the state-of-the-art methods regarding the mDice metric on the BraTS21 dataset}

\begin{tabular}{l|cccccc|c}
\hline\hline
\textbf{Methods} & \textbf{Client\textsubscript{1}} & \textbf{Client\textsubscript{2}} & \textbf{Client\textsubscript{3}} & \textbf{Client\textsubscript{4}} & \textbf{Client\textsubscript{5}} & \textbf{Client\textsubscript{6}} & \textbf{Average}   \\ \hline
FedAvg \cite{mcmahan2017communication}  & 25.68   & 46.57   & 53.72   & 48.91   & 65.61   & 65.46   & 50.99 \\ \hline
FedAAAI \cite{wu2024feda3i} & 41.88   & \textbf{52.6}   & 50.68   & 49.32   & 58.68   & 58.36   & 51.92 \\ \hline
FedProx \cite{li2020federated} & 35.90   & 48.55   & 56.16   & 45.79   & 64.32   & 65.11  & 52.64 \\ \hline
IOP-FL \cite{jiang2023iop}  & \textbf{48.39}   & 49.87   & 61.61   & \textbf{51.33}   & 47.93   & 63.08   & 53.70 \\ \hline
AAW \cite{pan2025adaptive} & 39.26   & 47.29   &60.23   & 43.61   &68.32   & \textbf{75.95}   & 55.78 \\ \hline
MDM-MixMFL (\textbf{ours}) & 46.26   & 45.24   & \textbf{67.61}   &47.29   & \textbf{74.05}   & 71.14   & \textbf{58.60} \\ \hline
iid (upper bound)    & {48.26}   & {46.06}   & {68.95}   & {47.93}   & {77.55}   & {68.58}   & {59.56 }\\
\hline\hline
\end{tabular}

\label{table1}
\end{table}

\begin{table}[]
\scriptsize
\centering
\caption{Comparison of state-of-the-art methods on the BraTS21 dataset under a three-modality-per-client MixMFL setting}

\begin{tabular}{l|cccccc|c}
\hline\hline
\textbf{Methods} &
\textbf{Client\textsubscript{1}} &
\textbf{Client\textsubscript{2}} &
\textbf{Client\textsubscript{3}} &
\textbf{Client\textsubscript{4}} &
\textbf{Average} \\ \hline

FedAvg \cite{mcmahan2017communication} &
37.75 &
48.73 &
52.08 &
51.68 &
47.56 \\ \hline

FedAAAI \cite{wu2024feda3i} &
35.50 &
46.48 &
62.20 &
68.92 &
53.28 \\ \hline

FedProx \cite{li2020federated} &
38.81 &
48.72 &
53.20 &
52.19 &
48.23 \\ \hline

IOP-FL \cite{jiang2023iop} &
\textbf{41.52} &
49.02 &
61.68 &
69.95 &
55.54 \\ \hline

AAW \cite{pan2025adaptive} &
38.03 &
47.56 &
63.92 &
\textbf{74.46} &
55.99 \\ \hline

MDM-MixMFL (\textbf{ours}) &
41.09 &
\textbf{49.63} &
\textbf{64.52} &
73.25 &
\textbf{57.12} \\

\hline\hline
\end{tabular}

\label{three-modality}
\end{table}

\subsection{Modality Memorizing}
The modality memorizing mechanism is explored to compensate for incomplete modalities in each local client by memorizing, refreshing, and retrieving modality prototypes, as shown in {Fig.}~\ref{Memory}. Through this mechanism, representative modality prototypes can be maintained and shared across clients during federated training. The retrieved prototypes are used to complement unavailable modalities at local clients, enabling more complete feature representations. This design enhances the robustness of the proposed framework under mix-modal federated learning scenarios. We allocate a memory bank with \(n\) slots for each modality, which is globally shared by all clients and is dynamically refreshed with the modality prototypes clustered from corresponding local modality-tailored representation.

During memorizing and refreshing procedures, the modality-tailored encoder in a local client will generate modality-specific representations for each sample during each local epoch. We perform lightweight clustering with these representations and obtain \( n \) cluster centers as corresponding modality prototypes. Taking the modality \( m \) in the \( t \)-th local epoch as an example, we will obtain the cluster centers \( \bm{P}^t_{m} = \{\bm{p}^t_{m,1},\bm{p}^t_{m,2}, ...,\bm{p}^t_{m,z} \}\). Where \(z\) is the number of cluster centers. Then, these cluster centers are written into the memory bank as modality prototypes. Compared to representative sample reserving measures, this memory mechanism avoids privacy exposure during federated aggregation. Obviously, the quality of modality prototypes will gradually improve with training iterations since the representation quality improves. Therefore, we adopt a First-in-first-out (FIFO) queue to write and refresh the modality memory. We denote the queue for modality \(m\) at epoch \(t\) as \(\bm{x}_m^{g,t}\).

During the retrieval procedure, we take the existing modality-specific representations as the semantic query to retrieve prototypes for incomplete modalities from the corresponding modality memory banks. Formally, assuming that a client  contains the modality \( a \) and the modality \( b \), each sample compensates for the incomplete modality \( c \) with a pseudo representation \(\hat{\bm{x}}_c^{g,t}\) by retrieving corresponding memory queue \(\bm{P}^t_c\) with existing semantics query \(\bm{x}_a^{g,t},\bm{x}_b^{g,t}\) via

\begin{align}
\hat{\bm{x}}_c^{g,t} = \mathop{\arg\max}_{\bm{q}_c} \ \ (sim(\bm{x}_a^{g,t},\bm{q}_c)+sim(\bm{x}_b^{g,t},\bm{q}_c))
\end{align}
Where $\bm{q}_c \in \bm{P}^t_{c}$, the function $ sim $ is a similarity calculation function (such as cosine similarity), \(\bm{x}_a^{g,t}\) and \(\bm{x}_b^{g,t}\) are the modality-specific representations generated from the existing modalities of each sample at the \(t\)-th epoch. Eventually, the modality-shared representation, the existing modality-tailored representation, and the compensated modality-tailored representation are concatenated and then fed into a globally shared decoder to generate the final segmentation.

\section{Experiments}

\begin{figure*}
    \centering
    \includegraphics[width=0.9\linewidth]{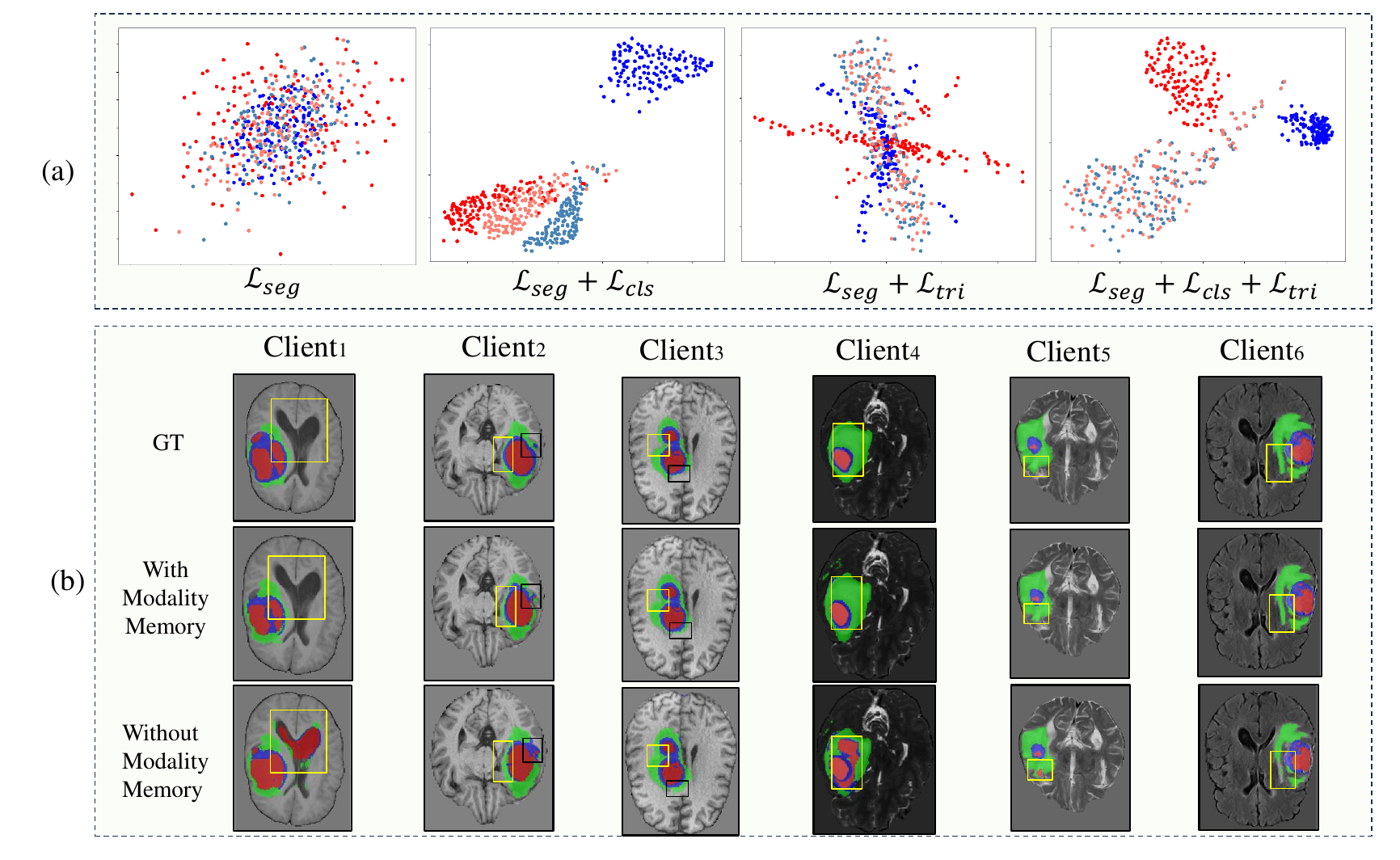}
    \caption{(a) Visualization of representation spaces guided by different variants. The colors red and blue denote two different modalities. The shade and dark dots of the same color (modality) denote representations generated from the modality-tailored encoder and modality-shared encoder, respectively. 
     (b) Visualization of segmentation results in different clients under conditions with and without the modality memory module, respectively.}
    \label{vision}
    % \vspace{0.35cm}
\end{figure*}

\subsection{Datasets}
\subsubsection{BraTS21} BraTS21 \cite{baid2021rsna} is a large-scale multimodal MRI brain glioma segmentation dataset with 1251 cases with pixel-level annotations, comprising 8,160 MRI scans from 2,040 patients. Each patient includes four modalities of MRI images: T1, T1ce, T2, and FLAIR. These images were acquired from multiple medical institutions using various clinical protocols and scanners. The annotations mainly include enhancing tumors (ET), peritumoral edema/invasive tissue (ED), and necrotic core of the tumor (NCR). The proportions of data that have these three types of annotations are 99.92\%, 96.56\%, and 97.36\%, respectively. All annotation labels and image data have been preprocessed, including aligning to a unified anatomical template, resampling to the same resolution (1 mm³), and skull stripping. We evenly distributed the dataset to 6 clients and divided each client's data into training and testing subsets with a ratio of 7:3.

\subsubsection{BraTS2023-MEN} BraTS2023-MEN \cite{labella2023asnr} dataset is a multimodal MRI brain meningiomas segmentation dataset that contains 1000 annotated training data, each providing four types of sequence MR input images (T1w, T1c, T2w, T2f) and segmentation results of meningiomas. The annotated content mainly includes non-enhanced tumor core (NETC), peripheral non-enhanced FLAIR high signal (SNFH), and enhanced tumor (ET). The proportions of data with the three types of annotations are 33.9\%, 53.6\%, and 99.9\%, respectively. We also split the dataset into training and testing subsets using the same protocol as the BraTS21 dataset.

\begin{figure*}
    \centering
    \includegraphics[width=0.9\linewidth]{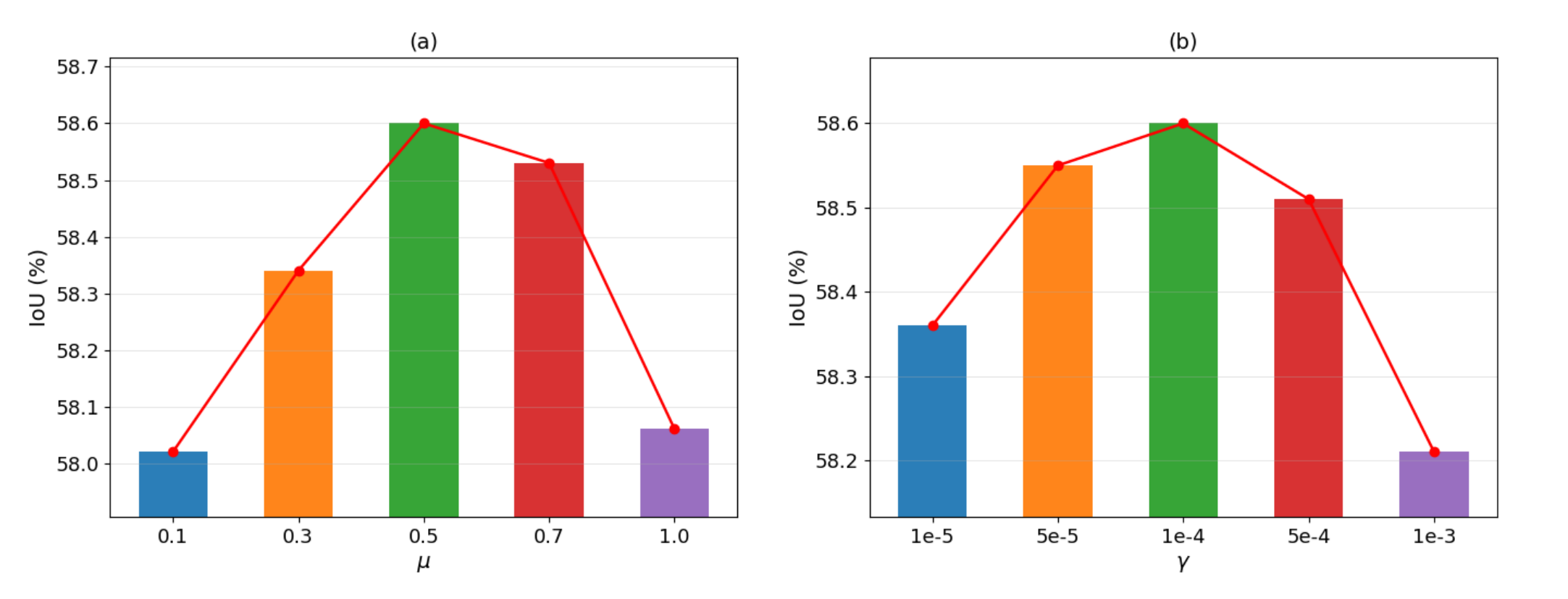}
    \caption{The influence of  classification loss weight $\mu$ and the triplet loss weight $\gamma$.}

    \label{line_chart}
    % \vspace{0.35cm}
\end{figure*}

\subsection{Implementation Details}
The above two datasets both include 4 modalities, we combine and distribute the four modalities to six clients (each with two modalities), where the modality combinations are all different across clients.  In addition, we include an extra experimental setting in which each client is equipped with three modalities, aiming to evaluate the generalizability of the proposed method under different modality-allocation configurations.  For each modality, we use three-fifths of the data for training, one-tenth of the data for validation, and the remaining data for testing. The data is equally divided into 6 parts and then distributed to 6 clients.

The proposed method is implemented with PyTorch on four NVIDIA RTX 3090 GPUs. We utilize TransBTS \cite{li2022transbtsv2} as our backbone. The input data of size 240×240×155 is randomly cropped to 128×128×128, and during both training and inference, thus a complete test image is split into 8 patches of size 128×128×128. These patches are independently used to obtain the segmentation results and then merged to form the whole full-resolution segmentation image. We use the Adam optimizer with a fixed batch size of 4. The initial learning rate is set to 0.0004 and is reduced by 0.1 during training. The training process stops after  200 epochs on the BraTS2021 dataset and BraTS2023-MEN dataset. Additionally, the memory size for each modality is set to 200 during each local epoch, and we use the classical k-means clustering algorithm to obtain cluster centers. The parameters \(m\) and \(\gamma\) are set to 0.5 and 0.0001, respectively.

\subsection{Comparison with State-of-the-art Methods}To validate the effectiveness and superiority of the proposed Modality decoupling and memorizing based mix-modal federated learning framework (MDM-MixMFL), we conduct comparison experiments on two large-scale heterogeneous distributed datasets for MRI image segmentation, including the BraTS21 and BraTS2023-MEN datasets. We report the mDice metric on six local clients and the average mDice across all clients for comprehensive performance comparison. All results are the averages of multiple runs under the same experimental setup. We use TransBTSV2 \cite{li2022transbtsv2} as the backbone and compare it with the following methods on the aforementioned two datasets: FedAVG \cite{mcmahan2017communication}, FedProx \cite{li2020federated}, FedAAAI \cite{wu2024feda3i}, IOP-FL \cite{jiang2023iop}, AAW\cite{pan2025adaptive}. Since there is no existing method under the proposed mix-modal federated learning (MixMFL) settings, we re-implemented these methods on the MixMFL settings based on their open-source code for fair comparisons.

\begin{table}[]
\scriptsize
\centering
\caption{Comparing with the state-of-the-art methods regarding the mDice metric on the BraTS2023-MEN dataset
}

\begin{tabular}{l|cccccc|c}
\hline\hline
\textbf{Methods} & \textbf{Client\textsubscript{1}} & \textbf{Client\textsubscript{2}} & \textbf{Client\textsubscript{3}} & \textbf{Client\textsubscript{4}} & \textbf{Client\textsubscript{5}} & \textbf{Client\textsubscript{6}} & \textbf{Average}   \\ \hline
FedAvg \cite{mcmahan2017communication}  & \textbf{39.48} & 48.29 & 39.04 & 52.12 & 30.70 & 22.74 & 38.73 \\ \hline
FedAAAI \cite{wu2024feda3i} & 37.21 & 46.95 & \textbf{42.32} & 49.23 & 37.2 & 22.51 & 39.24 \\ \hline
FedProx \cite{li2020federated} & 34.02 & 46.37 & 38.77 & 49.90 & 28.42 & 22.86 & 36.72 \\ \hline
IOP-FL \cite{jiang2023iop}  & 36.33 & 44.29 & 41.29 & 49.82 & \textbf{38.6} & 22.22 & 38.76 \\ \hline

AAW \cite{pan2025adaptive} &  34.18& \textbf{56.40} & 31.65 & 52.28 & 26.21 & 37.62 &39.72  \\ \hline
MDM-MixMFL (\textbf{ours}) & 31.80 & 55.25 & 31.85 & \textbf{55.49} & 31.22 &\textbf{40.54} & \textbf{41.03} \\ \hline\hline
\end{tabular}

\label{table2}
\end{table}

The segmentation results under the proposed MixMFL setting on the BraTS21 dataset are reported in {Table}~\ref{table1}. The proposed MDM-MixMFL framework shows a significant improvement regarding the mDice metric, achieving a performance increase of 2.82 (\%) compared to the second-best performing method. This improvement highlights the robustness of our approach in handling the challenges posed by modality heterogeneity and data heterogeneity.  As mentioned above, existing FL methods do not consider the mixed heterogeneity challenges induced by the MixMFL scenarios, and our MDM-MixMFL solves them by two measures, \ie, modality decoupling for personalized federated updating and modality memorizing for incomplete modality compensation. Specifically, our MDM-MixMFL overperforms the existing non-personalized federated learning methods (FedAVG \cite{mcmahan2017communication}, FedProx \cite{li2020federated}, FedAAAI \cite{wu2024feda3i}) by a large margin regarding the mDice metric. It is because they only seek a global-compromised but inevitably also a sub-optimal model for all clients. In contrast, the modality decoupling strategy in our MDM-MixMFL framework disentangles each modality into modality-tailored and modality-shared information, and then respectively tailored and shared updates corresponding modality encoders during federated aggregation, thus obtaining locally optimal tailored models for every client. Besides, our MDM-MixMFL also outperforms other personalized federated learning methods (IOP-FL \cite{jiang2023iop}) by a large margin of 4.9 (\%) regarding the mDice metric. We also include the AAW\cite{pan2025adaptive} method in our comparison, which is originally designed for multimodal federated learning without specific consideration of the proposed mix-modal setting. Our MDM-MixMFL achieves 58.60\% average mDice outperforming AAW\cite{pan2025adaptive} by +2.82\%. In addition, we report the upper bound of federated learning under the independent and identically distributed (i.i.d.) setting, and our MDM-MixMFL achieves an average performance only slightly lower than this upper bound. The superiority owes to two reasons: (1) Our MDM-MixMFL is equipped with a more fine-grained personalization strategy (\ie, the IOP-FL \cite{jiang2023iop} only personalizes at the model in each client while our modality decoupling strategy decouples each modality into more fine-grained modality-tailored and modality-shared parts. (2) The modality memorizing mechanism further compensates for incomplete modalities for each client during mix-modal federated updating. To further analyze the influence of loss weighting hyperparameters under the default two-modality-per-client BraTS21 setting, we conduct a sensitivity analysis on the classification loss weight $\mu$ and the triplet loss weight $\gamma$, as shown in Fig.~\ref{line_chart}. Specifically, we vary $\mu$ within $\{0.1, 0.3, 0.5, 0.7, 1.0\}$ and $\gamma$ within $\{10^{-5}, 5\times10^{-5}, 10^{-4}, 5\times10^{-4}, 10^{-3}\}$, while keeping all other settings unchanged. The results indicate that the adopted parameter configuration yields the best performance among the evaluated candidates.  In addition, Table~\ref{three-modality} reports the comparison results under a three-modality-per-client MixMFL setting on the BraTS21 dataset. The results indicate that the proposed MDM-MixMFL framework consistently achieves superior performance under this alternative modality-allocation configuration, suggesting that the proposed method remains effective and exhibits good robustness across different MixMFL settings.

Besides, the segmentation results under the proposed MixMFL setting on the BraTS2023-MEN dataset are reported in {Table}~\ref{table2}. We compare our MDM-MixMFL with four non-personalized methods (including the FedAVG \cite{mcmahan2017communication}, FedProx \cite{li2020federated}, and FedAAAI \cite{wu2024feda3i}), one personalized FL approach (\ie, IOP-FL \cite{jiang2023iop}), and one MulMFL approach (\ie, AAW \cite{pan2025adaptive}). Similarly, our MDM-MixMFL framework outperforms all state-of-the-art methods and outperforms the second-best approach with a large margin of 1.31 (\%) regarding the mDice coefficient. These result demonstrates the generalization and effectiveness of our method in a more challenging dataset since the BraTS2023-MEN dataset has more limited and unbalanced annotations (especially, there are only 339 samples that have been labeled in the NETC category).

\begin{table*}[]
\scriptsize
\centering
\caption{Ablation results of the MDM-MixMFL
 framework}

\begin{tabular}{l|cccccc|c}
\hline
\hline
{ \textbf{Removed Variants}}                                                                & { \textbf{client\textsubscript{1}}} & { \textbf{client\textsubscript{2}}} & { \textbf{client\textsubscript{3}}} & { \textbf{client\textsubscript{4}}} & { \textbf{client\textsubscript{5}}} & { \textbf{client\textsubscript{6}}} & { \textbf{Average}}   \\ \hline
{ \begin{tabular}[c]{@{}c@{}}No (Full Model)\\  \end{tabular}}                                                            & { 46.26}   & { 45.24}   & { 67.61}   & { 47.29}   & { 74.05}   & { 71.14}   & { \textbf{58.60}}\\ \hline
{ \begin{tabular}[c]{@{}c@{}} Tailored updating\end{tabular}}    & { 40.09}   & { 46.82}   & { 67.60}   & { 41.47}   & { 74.43}   & { 72.71}   & { 57.19 (-1.41)}\\ \hline
{ \begin{tabular}[c]{@{}c@{}}Modality memorizing\end{tabular}}              & { 43.93}   & { 40.42}   & { 68.26}   & { 43.69}   & { 74.06}   & { 71.79}   & { 57.14 (-1.46)}\\ \hline
{ \begin{tabular}[c]{@{}c@{}}Triple loss\end{tabular}}         & { 39.64}   & { 45.61}   & { 70.88}   & { 38.95}   & { 74.59}   & { 74.81}   & { 57.42 (-1.18)}\\ \hline
{ \begin{tabular}[c]{@{}c@{}}Classification loss\end{tabular}} & { 44.80}   & { 43.39}   & { 67.30}   & { 43.22}   & { 73.76}   & { 67.31}   & { 56.63 (-1.97)}\\ \hline
\hline
\end{tabular}

\label{table3}
\end{table*}

% \begin{table}[]
% \scriptsize
% \caption{Comparison with independent and identically distributed (iid) setting}
%  \resizebox{\columnwidth}{!}{
% \begin{tabular}{c|cccccc|c}
% \hline
% \hline
% { \textbf{Dataset settings}}                                                                & { \textbf{client\textsubscript{1}}} & { \textbf{client\textsubscript{2}}} & { \textbf{client\textsubscript{3}}} & { \textbf{client\textsubscript{4}}} & { \textbf{client\textsubscript{5}}} & { \textbf{client\textsubscript{6}}} & { \textbf{Average}}   \\ \hline
% { \begin{tabular}[c]{@{}c@{}}Mix-modal\\  \end{tabular}}                                                            & { 46.26}   & { 45.24}   & { 67.61}   & { 47.29}   & { 74.05}   & { 71.14}   & { 58.60}\\ \hline
% { \begin{tabular}[c]{@{}c@{}} iid\end{tabular}}    & { 48.26}   & { 46.06}   & { 68.95}   & { 47.93}   & { 77.55}   & { 68.58}   & { 59.56 }\\ \hline \hline             
% \end{tabular}
% }
% \label{table4}
% \end{table}

\subsection{Ablation Analysis}
The proposed MDM-MixMFL framework is mainly characterized by the modality decoupling strategy (which contains a classification loss and a triplet loss), and the modality memorizing mechanism. To examine the individual effectiveness of each component, we remove these components from our MDM-MixMFL framework respectively. As shown in {Table}~\ref{table3}, when removing the Tailored updating mechanism from our full model (MDM-MixMFL), the segmentation performance of the obtained variants degrades from 58.6(\%) to 57.19(\%) since it is impossible to decouple the unique information of each client. Without the modal memorizing mechanism, the segmentation performance of the obtained variants degrades from 58.6(\%) to 57.14(\%) since each client will not be able to compensate for its missing modal representation. Removing the Triple loss or Classification loss will respectively degrade the segmentation performance from 58.6(\%) to 57.42(\%) or 56.63(\%). This is because the triple loss and classification loss play a complementary role, and deleting either one will hinder the modality-decoupling process. In summary, deleting arbitrary modules or mechanisms in the proposed MDM-MixMFL will significantly reduce the final segmentation performance result, validating the individual effectiveness within our frameworks.

\subsection{Visualization Analysis}

To further intuitively illustrate the complementary and collaborative functions of the two branches in the modality decoupled. We exploit the MDS (Multidimensional Scaling) technique \cite{weinberger2006unsupervised} to visualize the representation space obtained from modality-tailored and modality-shared encoders under different auxiliary losses in {Fig.}~\ref{vision}~(a). We observe that applying classification loss alone could partially decouple modality-tailored and modality-shared representations but still leave some overlap, and also fail to align the shared components of two modalities generated from the modality-shared encoder. On the contrary, applying triplet loss alone could only bring the representations of two modalities generated from the modality-shared encoder closer (almost one-to-one correspondence) but fails to realize modality discrimination since the absence of classification loss. When both losses are applied, the modality-tailored and modality-shared representations are well-decoupled, and the modality-shared representations obtained from different modalities are also well-aligned.

To qualitatively validate the effectiveness of the modality compensation function of the modality memory module, we further visualize the brain tumor segmentation results under the conditions with and without the modality memory module, respectively. According to the previous studies \cite{ahamed2023review} for segmenting brain tumors in MRI images, the T1 and  T1ce modalities are usually good for highlighting the tumor core (the red area in {Fig.}~\ref{vision}~(b)), while the T2 and FLAIR are good for highlighting peritumoral edema (the blue area in {Fig.}~\ref{vision}~(b)). As shown in {Fig.}~\ref{vision}~(b), Client3 (who only holds T1 and T1ce modalities) performs significantly better in segmenting peritumoral edema (green) with a modality memory module than without it. Similarly, Client4 (who only holds T2 and FLAIR modalities) also performs significantly better in segmenting the tumor core (red) with a modality memory module than without it. These results clearly indicate the modality compensating effectiveness from the proposed modality memory module during mix-modal federated learning.

\section{Conclusion}
In this paper, we successfully formulate the non-centralized mix-modal MRI image segmentation as a mix-modal federating learning (MixMFL) paradigm, which defines a new paradigm of federating learning that focuses on mixed heterogeneity both from data and modality. Then, we propose a novel modality decoupling and memorizing (MDM) based framework for this paradigm. Specifically, its modality decoupling strategy disentangles modality-tailored and shared information, ensuring stable and adaptive federated updating. Additionally, the modality memorizing mechanism bridges modality gaps by dynamically memorizing and refreshing the modality prototypes from each modality-tailored encoder, compensating for incomplete modality data in local clients. This mechanism enhances modality aggregation and fusion during federated learning, further improving the final segmentation Performance. Extensive experiments on two public MRI segmentation datasets demonstrate that MDM-MixMFL outperforms state-of-the-art methods, highlighting its effectiveness and robustness in addressing the mixed challenges of data heterogeneity and modality heterogeneity in practical decentralized medical scenarios.

\noindent\textbf{Acknowledgments}: This work was supported in part by the National Natural Science Foundation of China (No. 62506003, No. 62376004, No. U24A20342), in part by the Anhui Provincial Natural Science Foundation (No. 2408085QF201), and in part by the Open Project of Anhui Provincial Key Laboratory of Intelligent Detection and Diagnosis for Traffic Infrastructure (No. KY-2025-03).

\bibliographystyle{ieeetr}
\bibliography{ref}

@ARTICAL{ahamed2023review,
  title={A review on brain tumor segmentation based on deep learning methods with federated learning techniques},
  author={Ahamed, Md Faysal and Hossain, Md Munawar and Nahiduzzaman, Md and Islam, Md Rabiul and Islam, Md Robiul and Ahsan, Mominul and Haider, Julfikar},
  journal={Computerized Medical Imaging and Graphics},
  pages={102313},
  year={2023},
  publisher={Elsevier}
}

@article{yang2024cross,
  title={Cross-Modal Federated Human Activity Recognition},
  author={Yang, Xiaoshan and Xiong, Baochen and Huang, Yi and Xu, Changsheng},
  journal={IEEE Transactions on Pattern Analysis and Machine Intelligence},
  year={2024},
  publisher={IEEE}
}

@inproceedings{dai2024federated,
  title={Federated Modality-Specific Encoders and Multimodal Anchors for Personalized Brain Tumor Segmentation},
  author={Dai, Qian and Wei, Dong and Liu, Hong and Sun, Jinghan and Wang, Liansheng and Zheng, Yefeng},
  booktitle={Proceedings of the AAAI Conference on Artificial Intelligence},
  volume={38},
  number={2},
  pages={1445--1453},
  year={2024}
}

@inproceedings{ouyang2024admarker,
  title={ADMarker: A Multi-Modal Federated Learning System for Monitoring Digital Biomarkers of Alzheimer's Disease},
  author={Ouyang, Xiaomin and Shuai, Xian and Li, Yang and Pan, Li and Zhang, Xifan and Fu, Heming and Cheng, Sitong and Wang, Xinyan and Cao, Shihua and Xin, Jiang and others},
  booktitle={Proceedings of the 30th Annual International Conference on Mobile Computing and Networking},
  pages={404--419},
  year={2024}
}

@incollection{wong2005medical,
  title={Medical image segmentation: methods and applications in functional imaging},
  author={Wong, Koon-Pong},
  booktitle={Handbook of Biomedical Image Analysis: Volume II: Segmentation Models Part B},
  pages={111--182},
  year={2005},
  publisher={Springer}
}

@inproceedings{zeineldin2022multimodal,
  title={Multimodal CNN networks for brain tumor segmentation in MRI: a BraTS 2022 challenge solution},
  author={Zeineldin, Ramy A and Karar, Mohamed E and Burgert, Oliver and Mathis-Ullrich, Franziska},
  booktitle={International MICCAI Brainlesion Workshop},
  pages={127--137},
  year={2022},
  organization={Springer}
}

@article{ottom2022znet,
  title={Znet: deep learning approach for 2D MRI brain tumor segmentation},
  author={Ottom, Mohammad Ashraf and Rahman, Hanif Abdul and Dinov, Ivo D},
  journal={IEEE Journal of Translational Engineering in Health and Medicine},
  volume={10},
  pages={1--8},
  year={2022},
  publisher={IEEE}
}

@article{sun2019brain,
  title={Brain tumor segmentation and survival prediction using multimodal MRI scans with deep learning},
  author={Sun, Li and Zhang, Songtao and Chen, Hang and Luo, Lin},
  journal={Frontiers in neuroscience},
  volume={13},
  pages={810},
  year={2019},
  publisher={Frontiers Media SA}
}

@article{shi2023m,
  title={MFTrans: Modality-Masked Fusion Transformer for Incomplete Multi-Modality Brain Tumor Segmentation},
  author={Shi, Junjie and Yu, Li and Cheng, Qimin and Yang, Xin and Cheng, Kwang-Ting and Yan, Zengqiang},
  journal={IEEE Journal of Biomedical and Health Informatics},
  year={2023},
  publisher={IEEE}
}

@article{liu2024medmap,
  title={MedMAP: Promoting Incomplete Multi-modal Brain Tumor Segmentation with Alignment},
  author={Liu, Tianyi and Tan, Zhaorui and Chen, Muyin and Yang, Xi and Jiang, Haochuan and Huang, Kaizhu},
  journal={arXiv preprint arXiv:2408.09465},
  year={2024}
}

@article{xiong2022unified,
  title={A unified framework for multi-modal federated learning},
  author={Xiong, Baochen and Yang, Xiaoshan and Qi, Fan and Xu, Changsheng},
  journal={Neurocomputing},
  volume={480},
  pages={110--118},
  year={2022},
  publisher={Elsevier}
}

@inproceedings{chen2022towards,
  title={Towards optimal multi-modal federated learning on non-IID data with hierarchical gradient blending},
  author={Chen, Sijia and Li, Baochun},
  booktitle={IEEE INFOCOM 2022-IEEE conference on computer communications},
  pages={1469--1478},
  year={2022},
  organization={IEEE}
}

@article{li2024feddiff,
  title={FedDiff: Diffusion model driven federated learning for multi-modal and multi-clients},
  author={Li, Daixun and Xie, Weiying and Wang, Zixuan and Lu, Yibing and Li, Yunsong and Fang, Leyuan},
  journal={IEEE Transactions on Circuits and Systems for Video Technology},
  year={2024},
  publisher={IEEE}
}

@article{yu2024robust,
  title={Robust multimodal federated learning for incomplete modalities},
  author={Yu, Songcan and Wang, Junbo and Hussein, Walid and Hung, Patrick CK},
  journal={Computer Communications},
  volume={214},
  pages={234--243},
  year={2024},
  publisher={Elsevier}
}

@article{labella2023asnr,
  title={The asnr-miccai brain tumor segmentation (brats) challenge 2023: Intracranial meningioma},
  author={LaBella, Dominic and Adewole, Maruf and Alonso-Basanta, Michelle and Altes, Talissa and Anwar, Syed Muhammad and Baid, Ujjwal and Bergquist, Timothy and Bhalerao, Radhika and Chen, Sully and Chung, Verena and others},
  journal={arXiv preprint arXiv:2305.07642},
  year={2023}
}

@article{li2022transbtsv2,
  title={TransBTSV2: towards better and more efficient volumetric segmentation of medical images},
  author={Li, Jiangyun and Wang, Wenxuan and Chen, Chen and Zhang, Tianxiang and Zha, Sen and Wang, Jing and Yu, Hong},
  journal={arXiv preprint arXiv:2201.12785},
  year={2022}
}

@inproceedings{mcmahan2017communication,
  title={Communication-efficient learning of deep networks from decentralized data},
  author={McMahan, Brendan and Moore, Eider and Ramage, Daniel and Hampson, Seth and y Arcas, Blaise Aguera},
  booktitle={Artificial intelligence and statistics},
  pages={1273--1282},
  year={2017},
  organization={PMLR}
}

@article{li2020federated,
  title={Federated optimization in heterogeneous networks},
  author={Li, Tian and Sahu, Anit Kumar and Zaheer, Manzil and Sanjabi, Maziar and Talwalkar, Ameet and Smith, Virginia},
  journal={Proceedings of Machine learning and systems},
  volume={2},
  pages={429--450},
  year={2020}
}

@inproceedings{wu2024feda3i,
  title={FedA3I: Annotation Quality-Aware Aggregation for Federated Medical Image Segmentation against Heterogeneous Annotation Noise},
  author={Wu, Nannan and Sun, Zhaobin and Yan, Zengqiang and Yu, Li},
  booktitle={Proceedings of the AAAI Conference on Artificial Intelligence},
  volume={38},
  number={14},
  pages={15943--15951},
  year={2024}
}

@article{jiang2023iop,
  title={IOP-FL: Inside-outside personalization for federated medical image segmentation},
  author={Jiang, Meirui and Yang, Hongzheng and Cheng, Chen and Dou, Qi},
  journal={IEEE Transactions on Medical Imaging},
  volume={42},
  number={7},
  pages={2106--2117},
  year={2023},
  publisher={IEEE}
}

@article{shwartz2017opening,
  title={Opening the black box of deep neural networks via information},
  author={Shwartz-Ziv, Ravid and Tishby, Naftali},
  journal={arXiv preprint arXiv:1703.00810},
  year={2017}
}

@article{baid2021rsna,
  title={The rsna-asnr-miccai brats 2021 benchmark on brain tumor segmentation and radiogenomic classification},
  author={Baid, Ujjwal and Ghodasara, Satyam and Mohan, Suyash and Bilello, Michel and Calabrese, Evan and Colak, Errol and Farahani, Keyvan and Kalpathy-Cramer, Jayashree and Kitamura, Felipe C and Pati, Sarthak and others},
  journal={arXiv preprint arXiv:2107.02314},
  year={2021}
}

@inproceedings{milletari2016v,
  title={V-net: Fully convolutional neural networks for volumetric medical image segmentation},
  author={Milletari, Fausto and Navab, Nassir and Ahmadi, Seyed-Ahmad},
  booktitle={2016 fourth international conference on 3D vision (3DV)},
  pages={565--571},
  year={2016},
  organization={Ieee}
}

@article{weinberger2006unsupervised,
  title={Unsupervised learning of image manifolds by semidefinite programming},
  author={Weinberger, Kilian Q and Saul, Lawrence K},
  journal={International journal of computer vision},
  volume={70},
  pages={77--90},
  year={2006},
  publisher={Springer}
}

@article{huang2022eefed,
  title={EEFED: Personalized federated learning of execution\&evaluation dual network for CPS intrusion detection},
  author={Huang, Xianting and Liu, Jing and Lai, Yingxu and Mao, Beifeng and Lyu, Hongshuo},
  journal={IEEE Transactions on Information Forensics and Security},
  volume={18},
  pages={41--56},
  year={2022},
  publisher={IEEE}
}

@article{huang2024federated,
  title={Federated Learning with Long-Tailed Data via Representation Unification and Classifier Rectification},
  author={Huang, Wenke and Liu, Yuxia and Ye, Mang and Chen, Jun and Du, Bo},
  journal={IEEE Transactions on Information Forensics and Security},
  year={2024},
  publisher={IEEE}
}

@article{zhou2022pflf,
  title={PFLF: Privacy-preserving federated learning framework for edge computing},
  author={Zhou, Hao and Yang, Geng and Dai, Hua and Liu, Guoxiu},
  journal={IEEE Transactions on Information Forensics and Security},
  volume={17},
  pages={1905--1918},
  year={2022},
  publisher={IEEE}
}

@article{wang2025interpretable,
  title={Interpretable vertical federated learning with privacy-preserving multi-source data integration for prognostic prediction},
  author={Wang, Qingyong},
  journal={Engineering Applications of Artificial Intelligence},
  volume={148},
  pages={110408},
  year={2025},
  publisher={Elsevier}
}

@article{zhang2024afl,
  title={AFL-DCS: An asynchronous federated learning framework with dynamic client scheduling},
  author={Zhang, Ruizhuo and Luo, Wenjian and Luo, Yongkang and Zhang, Hongwei and Wang, Jiahai},
  journal={Engineering Applications of Artificial Intelligence},
  volume={133},
  pages={107927},
  year={2024},
  publisher={Elsevier}
}

@inproceedings{pan2025adaptive,
  title={Adaptive Aggregation Weights for Federated Segmentation of Pancreas MRI},
  author={Pan, Hongyi and Durak, Gorkem and Zhang, Zheyuan and Taktak, Yavuz and Keles, Elif and Aktas, Halil Ertugrul and Medetalibeyoglu, Alpay and Velichko, Yury and Spampinato, Concetto and Schoots, Ivo and others},
  booktitle={2025 IEEE 22nd International Symposium on Biomedical Imaging (ISBI)},
  pages={1--5},
  year={2025},
  organization={IEEE}
}

@article{hu2025federated,
  title={Federated Client-tailored Adapter for Medical Image Segmentation},
  author={Hu, Guyue and Song, Siyuan and Kang, Yukun and Yin, Zhu and Zhao, Gangming and Li, Chenglong and Tang, Jin},
  journal={IEEE Transactions on Information Forensics and Security},
  year={2025},
  publisher={IEEE}
}

@article{le2025cross,
  title={Cross-modal prototype based multimodal federated learning under severely missing modality},
  author={Le, Huy Q and Thwal, Chu Myaet and Qiao, Yu and Tun, Ye Lin and Nguyen, Minh NH and Huh, Eui-Nam and Hong, Choong Seon},
  journal={Information Fusion},
  pages={103219},
  year={2025},
  publisher={Elsevier}
}

@article{orzikulova2024federated,
  title={Federated learning with incomplete sensing modalities},
  author={Orzikulova, Adiba and Kwak, Jae Hyun and Shin, Jaemin and Lee, Sung-Ju},
  journal={CoRR},
  year={2024}
}

@article{li2025ckdf,
  title={CKDF-V2: Effectively Alleviating Representation Shift for Continual Learning With Small Memory},
  author={Li, Kunchi and Chen, Hongyang and Wan, Jun and Yu, Shan},
  journal={IEEE Transactions on Neural Networks and Learning Systems},
  year={2025},
  publisher={IEEE}
}

@article{li2025enhance,
  title={Enhance the old representations’ adaptability dynamically for exemplar-free continual learning},
  author={Li, Kunchi and Ding, Chaoyue and Wan, Jun and Yu, Shan},
  journal={Neurocomputing},
  pages={130286},
  year={2025},
  publisher={Elsevier}
}

@article{pan2025diverse,
  title={Diverse feature generation for zero-shot chinese character recognition},
  author={Pan, Song-Liang and Li, Kunchi and Wang, Da-Han and Zhang, Xu-Yao and Liu, Jiantao and Zhu, Shunzhi},
  journal={Expert Systems with Applications},
  pages={129442},
  year={2025},
  publisher={Elsevier}
}

@article{hu2025dynamic,
  title={Dynamic strip convolution and adaptive morphology perception plugin for medical anatomy segmentation},
  author={Hu, Guyue and Kang, Yukun and Zhao, Gangming and Jin, Zhe and Li, Chenglong and Tang, Jin},
  journal={IEEE Transactions on Medical Imaging},
  year={2025},
  publisher={IEEE}
}

@article{hu2025missingness,
  title={Missingness-aware prompting for modality-missing RGBT tracking},
  author={Hu, Guyue and Wang, Zhanghuan and Li, Chenglong and Yuan, Duzhi and He, Bin and Tang, Jin},
  journal={Journal of King Saud University Computer and Information Sciences},
  volume={37},
  number={6},
  pages={128},
  year={2025},
  publisher={Springer}
}

\end{document}